\documentclass[11pt]{article}

\usepackage[preprint]{acl}

\usepackage{times}
\usepackage{latexsym}

\usepackage{dblfloatfix}

\usepackage[T1]{fontenc}
\usepackage[utf8]{inputenc}

\usepackage{microtype}

\usepackage{inconsolata}

\usepackage{graphicx}
\usepackage{enumitem}

\usepackage{amsmath}
\usepackage{amssymb}
\usepackage{placeins}

\title{Human-Alignment, Calibration, and Activation Patterns\\in Large Language Model Uncertainty}

\author{
    \textbf{Kyle Moore\textsuperscript{1}},
    \textbf{Jesse Roberts\textsuperscript{2}},
    \textbf{Daryl Watson\textsuperscript{2}},
    \textbf{William Ward\textsuperscript{2}},
    \textbf{Grayson Heyboer\textsuperscript{2}}
  \\
  \\
 \textsuperscript{1}Vanderbilt University \\
 \textsuperscript{2}Tennessee Technological University
 \\
 \small{
   \textbf{Correspondence:} \href{mailto:kyle.a.moore@vanderbilt.edu}{kyle.a.moore@vanderbilt.edu} / \href{mailto:jtroberts@tntech.edu}{jtroberts@tntech.edu}
 }
  }

\begin{document}
\maketitle
\begin{abstract}
Uncertainty Quantification is a large and growing subfield of large language model behavioral analysis. Primarily to recognize and combat hallucination, the field has largely focused on measuring and improving calibration, the accuracy of uncertainty judgments to task efficacy. In this work, we investigate the relatively underexplored question of how similar large language model uncertainty is to human uncertainty. We investigate the presence and strength of human-similar uncertainty signals, deemed uncertainty alignment, in large language model overt behavior and internal activation patterns. We identify whether the models show evidence of simultaneous alignment and calibration on a variety of datasets covering both multiple choice and open ended factual recall. And we characterize the effect of instruct fine-tuning on each of these facets.
\renewcommand{\thefootnote}{}   % empty mark
\footnote{All code available under permissive license at \url{https://github.com/KyleAMoore/LLM-UQ-Align-and-Calibrate}}
\renewcommand{\thefootnote}{\arabic{footnote}}  % restore
\addtocounter{footnote}{-1}

\end{abstract}

\section{Introduction}

Quantification and characterization of the uncertainty of transformer-based large language models (LLMs) has emerged as a major research focus. Accurate uncertainty quantification (UQ) is an important factor in the recognition and mitigation of model hallucinations \citep{farquhar2024detecting} and the building and maintenance of user trust in LLM-based applications under threat of hallucination. While much research has been dedicated to the development of UQ methods and models that are well-calibrated, meaning that the uncertainty accurately indicates model efficacy \citep{guo2017calibration, shorinwa2025survey}, very little work has investigated the related question of whether LLMs are human-like in their uncertainty behavior.

We believe the question of human-similarity in LLMs may have significant implications for human-AI collaboration. Greater levels of similarity in how uncertainty occurs and presents may foster greater rapport between LLMs and users when uncertainty judgments are provided, as human users might safely transfer their intuitive expectations of behavior under uncertainty from human-human interpersonal interaction to human-AI collaboration. Conversely, dissimilar uncertainty incidence may benefit human-AI collaboration by allowing each to cover for the other's weaknesses. 

Beyond strict utility, human-similarity (or dissimilarity) in LLM uncertainty may have significant implications for the transferability of human research on behavior under uncertainty to expectations of AI behavior and vice versa. Both similarity and dissimilarity may also yield further questions about the mechanisms that underlie either, including comparisons in memory organization and information availability during learning.

% A growing body of work has investigated the presence of human-like behavioral phenomena that are, in humans, modulated by uncertainty, including fan effects \citep{roberts2024large}, choice-preference effects \citep{moore2024base}, and various constrained reasoning heuristics \citep{various works that look at kahneman-esque heuristics}. To our knowledge, only one work to date has investigated whether uncertainty varies contextually similarly to humans \citep{Moore_Roberts_Watson_Wisniewski_2026}. They defined this property, which they dubbed uncertainty alignment, as the correlation across a shared set of task instances between human uncertainty level and LLM uncertainty level. Despite significant limitations in task coverage and dataset size/diversity, they found positive evidence of a weak uncertainty alignment signal.

In this work, we expand on the existing literature in both breadth and depth to better understand the incidence, presentation, and source of human aligned uncertainty across a diverse variety of LLMs. We start by expanding the work of \citet{Moore_Roberts_Watson_Wisniewski_2026} to a larger breadth of model coverage and a set of larger and more diverse datasets. We make a first foray into measuring uncertainty alignment in a free response context. Uncertainty alignment and calibration are then tied together by measuring simultaneous alignment and calibration in identical contexts. Finally, we probe LLM activations to characterize how and where uncertainty alignment arises internally. We identify, in short, the following primary findings:

\begin{enumerate}[noitemsep,topsep=2pt]
    \item LLM uncertainty is generally weakly-to-moderately positively correlated with human uncertainty, mediated by task and model.
    \item Both uncertainty alignment and calibration are significantly degraded by instruct-finetuning.
    \item Human-aligned uncertainty is more strongly detectable in internal activations than in output logits, with qualitative differences between group-level and individual-level alignment.
\end{enumerate}

\section{Uncertainty Quantification}
Uncertainty quantification in LLMs has been extensively studied, typically with the stated goal of detecting and reducing model hallucination. In this section, we briefly survey the existing UQ research and how calibration, the typical target of UQ research, is measured.

\subsection{Prior Work}
Uncertainty quantification in the general sense has grown rapidly as a subfield of LLM research. The techniques employed have grown numerous and have been well-surveyed \citep{shorinwa2025survey, liu2025uncertainty}. Research in this field is typically targeted towards identifying and fostering uncertainty calibration, with relatively little work focusing on establishing human similarity. Existing work has investigated how similar the calibration of LLMs is to that of humans \citep{sun2025large}, how human estimations of LLM certainty differ from true uncertainty \citep{steyvers2025large}, and whether uncertainty-modulated human-like behaviors are also exhibited by LLMs \citep{xu2025language, roberts2024large, moore2025chain}. To our knowledge, the only existing work that investigates uncertainty alignment directly is \citet{Moore_Roberts_Watson_Wisniewski_2026}, which found initial but weak evidence of uncertainty alignment in LLaMa and Mistral models using a small subset of the measures we here employ. Additionally, they did not investigate model activations or open response.

\subsection{Expected Calibration Error}
Model calibration is most commonly measured in terms of expected calibration error (ECE), defined as the expected absolute difference between a model's confidence (generally taken as the additive inverse of uncertainty) and the model's accuracy given that confidence level \citep{guo2017calibration}. 

Because model correctness is binary per instance, this measure is typically estimated using a binned approximation. Each of the $n$ instances are separated into a set of $m$ bins based on confidence level, giving each bin an associated within-bin accuracy, $acc(B_i)$, and within-bin average confidence, $conf(B_i)$. The ECE is then taken to be the weighted average absolute difference between these two quantities across all bins (Equation \ref{eq:ece-binned}).

\begin{equation}
    \small
    \label{eq:ece-binned}
    ECE \approx \sum_{m=1}^M \frac{|B_m|}{n}|acc(B_m) - conf(B)m)|
\end{equation}

ECE is known to be sensitive to the choice of binning procedure with both the number of bins and the across-bin probability mass distribution impacting how successfully ECE captures true calibration \citep{wang2023calibration}. In this work, all LLM calibration measurements use a technique called ECESweep \citep{roelofs2022mitigating}. This method dynamically determines bin count such that the number of bins, $m$, is maximized while maintaining monotonicity of $acc(B_i)$ with increasing $i$ for all bin counts in the range $[2,m]$. The authors find that this method effectively reduces binning bias and better detects miscalibration.

\section{Methods}
In this section, we describe the experimental details, covering the variety of LLMs tested (Section \ref{sec:methods-models}), the datasets used (Section \ref{sec:methods-datasets}), the methods used for quantifying LLM uncertainty (Section \ref{sec:methods-uq-measures}), model inference and query construction (Section \ref{sec:methods-inference}), and activation probing techniques employed (Section \ref{sec:methods-probing}).

\subsection{Model Choice}
\label{sec:methods-models}
We use $30$ open-weight text-only LLMs representing a variety of model families (covering LLaMa, Mistral, Gemma, and Falcon) and models sizes (ranging from 1B to 13B parameters per model). Specifically, we tested LLaMa 2 (7B and 13B) \citep{touvron2023llama}, LLaMa 3 (1B, 3B, and 8B) \citep{grattafiori2024llama}, Mistral (0.1 and 0.3, both 7B) \citep{jiang2023mistral7b}, Gemma (2B and 7B) \citep{team2024gemma}, Gemma (2B and 9B) \citep{team2024gemma2}, and Falcon 3 (1B, 3B, 7B, and 10B) \citep{Falcon3}. In all cases, we tested both the instruct-finetuned and base versions of every model.

\subsection{Datasets}
\label{sec:methods-datasets}
All non-probing experiments were performed identically over four datasets. All datasets are question answering tasks with substantial variation in topic coverage and question format.

Our first dataset is the commonly employed multiple choice question answering (MCQA) MMLU dataset \citep{hendrycks2020measuring}. This dataset consists of more than $14K$ multiple choice questions drawn from a wide variety of factual subjects. As this dataset provides no information regarding per-question human uncertainty response, it is included primarily as an established baseline when measuring ECE for the remaining datasets.

Each of the remaining datasets provide human response data from which we derive estimates of per-question human uncertainty. The first is ProtoQA \citep{boratko2020protoqa}, which is a Family Feud inspired dataset consisting of 8724 crowd-sourced purely preference or opinion-based questions. Each question has a variable-size set of answers provided by human survey respondents. On average, each question is associated with $5.033$ answer choices and $89.407$ independent human respondents. The second dataset, CamChoice \citep{mullooly2023cambridge} comprises $504$ reading comprehension questions where each question is associated with a passage on which the question is based. All questions in this dataset have exactly $4$ available answer choices and each passage is associated with an average of $6.46$ questions. During inference, passages are prepended to questions such that each question and its associated passage are presented to the model in isolation. For both the CamChoice and ProtoQA datasets, we have access to the percentage of human respondents that selected each of the available answer choices. For these datasets, we took the human uncertainty to be the normalized entropy over these response percentages, which we consider a measure of \textit{group uncertainty}.

The final dataset is named herein the Coane dataset, after the first author of \citet{coane2021database}. This dataset consists of $421$ general knowledge questions that were presented to older adult subjects. Like ProtoQA and CamChoice, the Coane dataset provides the percent of subjects who selected each of $4$ answer choices per question, allowing the computation of entropy over these responses as well. In addition, Coane also provides the average response time. They separate response times based on correctness, but provide sufficient information to reconstruct the average response time across all subjects, which we consider a measure of \textit{average individual uncertainty} following the established interpretation in cognitive science \cite{hick1952rate}. In addition to the MCQA task, Coane also provides human response data on an open ended question answering (OEQA) variant over the same question set. This allows us to investigate uncertainty alignment in both MCQA and OEQA contexts without substantial differences in context. The human data provided allows us to calculate the average response time over all respondents, which we interpret as the average individual uncertainty in a OEQA setting.

\subsection{Uncertainty Measures}
\label{sec:methods-uq-measures}
Our experiments are limited to a wide selection of inference-time LLM UQ measures that can be calculated from the instantaneous model logits and activations at each inference timestep. Such per-iteration uncertainty signals, in addition to being computationally cheaper than multi-inference UQ methods, are more conducive to signal utilization in downstrean applications and analysis. For example, future analysis regarding uncertainty incidence throughout generation would be difficult or impossible with only access to sequence-level uncertainty values. To facilitate broad coverage under possibility of measure choice-senstivity and coverage of both MCQA and OEQA tasks, we experiment with 4 broad but interconnected categories of UQ measures: MCQA measures, Free Response (FR), Bracketed FR (BFR), and 1st token FR (1TFR).

\subsubsection{MCQA measures}
The basic set of uncertainty measures are defined for MCQA and are largely adapted from those used in \citet{Moore_Roberts_Watson_Wisniewski_2026}. These measures fall into 5 subcategories: full distribution measures, cloze measures, top-k entropies, top-p entropies, and top-p sizes. There are two full distribution measures: top-1-prob and total-ent. The former is simply the additive inverse of the highest probability token across the entire vocabulary, that is $1-\max_{v\in V}(P(v|C_{MC})$ where $C_{MC}$ is a multiple-choice formatted context and $V$ is the full model vocabulary. The additive inverse is taken because higher probability mass on the top token suggests higher certainty, therefore lower uncertainty. The other measure, total-ent, is the normalized entropy over the full vocabulary set (Equation \ref{eq:total-ent}). As probability mass becomes more diffuse across the vocabulary, suggesting that either more tokens are taken as valid alternatives or the most preferred token(s) become less clearly preferred, the entropy of the distribution will rise. This intuitively suggests entropy over outcomes (vocabulary tokens) as a valid measure of model uncertainty. This is established in the literature \citep{huang2025look}, but is relatively uncommonly used in favor of more complex methods.

\begin{equation}
    -\frac{1}{\ln|V|}\sum_{v\in V}P(v|C_{MC})\ln(P(v|C_{MC})
    \label{eq:total-ent}
\end{equation}

All remaining measures are calculated based on various focused subsets of the total probability distribution. These are chosen based on expectation that the probability mass and the meaningful uncertainty signal will typically be concentrated in a small portion of the overall vocabulary. Focusing exclusively on these vocabulary subsets may allow identification of signals that would otherwise be lost in the full distribution measures or better discernment between uncertainty and low entropy caused by task-irrelevant syntactic or semantic restriction of the probable token set. Most of these remaining measures are defined as the entropy over three focus subsets: cloze, top-k, and top-p. In all cases, the entropy is calculated exactly as in Equation \ref{eq:total-ent} with two major changes.

First, the vocabulary, $V$, is replaced by a strict subset $V_k$, $V_p$, or $V_c$. $V_k$ represents the $k$ highest probability tokens from $P(V|C_{MC})$. $V_p$ is the top-p tokens, meaning the minimum set of tokens such that $\sum_{v\in V_p}P(v)\geq p$. $V_c$ is a pre-defined set of tokens that represent the set of valid cloze test completions for the MCQA context. 

To handle semantically equivalent variations, probabilities are summed across variations of valid cloze tokens. For example, the probability of the cloze target `A' is $P(A\in V_c) = P(\text{`}A\text{'}) + P(\text{`}A)\text{'})+P(\text{`}A.\text{'})+P(\text{`}a\text{'})+\dots)$. The probability of the tokens within the subsets are renormalized to sum to $1$ by dividing by the total subset probability, that is $P(v^i\in V_i) = P(v\in V) / \big(\sum_{v\in V_i}P(v)\big)$. 

For top-k and top-p entropies, uncertainty is measured across a range of reasonable subset sizes, specifically $k\in\{5,10,25,50,100\}$ and $p\in\{0.95,0.90,0.75,0.50\}$.

The remaining measures are not entropy based. The first is the top-1 probability over the cloze token set, $V_c$ (after the previously described semantic grouping). Ideally, but not necessarily, the most likely token given an MCQA-formatted context will be one of the valid tokens within the cloze token target set. This yields an expectation, but not a guarantee, that top-1-prob is less than but nearly equal or proportional to choice-prob. The remaining measures are derived from the top-p subset, $V_p$. We take the number of tokens that fall into this set $|V_p|$ to be a measure of uncertainty intuitively representing the number of valid completion tokens given $C_{MC}$. This measure is alternatively known in the literature as the nucleus size and is connected to the credible interval, which is the Bayesian analog of the commonly employed confidence interval.

\subsubsection{FR, BFR, and 1TFR Measures}
The various free response uncertainty measures are largely derived directly from the MC measures described in the previous section. The only significant exception to this is sequence perplexity (PPL). Perplexity, defined as used herein in Equation \ref{eq:ppl} where $N$ is the number of generated tokens and $W$ is the generated token sequence, is a common metric of sequence probability and surprisal. While PPL is typically defined over the entire sequence, including both starting context and model response, we take the perplexity over only the generated sequence to avoid confounding effects from relative likelihood across varying contexts. Being analogous to sequence-level probability, this measure was also converted to uncertainty by taking its additive inverse, $1-PPL(W|C_{FR})$.

\begin{align}
    PPL&(W|C_{FR}) = \nonumber\\
    &exp(-\frac{1}{N}\sum_{i=1}^N\ln P(w_i | C_{FR},w_1,\dots,w_{i-1}))
    \label{eq:ppl}
\end{align}

The remaining FR measures are simple averages over all generation steps of the non-cloze measures as defined in the previous section. This includes the generated sequence average total entropy, top-k entropy, top-p entropy, and top-p size. The entropy measures deserve some additional justification for this choice. Entropy over the probability distribution is defined as the expected log surprisal of each outcome. Because the resulting value is therefore in logspace, the arithmetic mean of entropy is equivalent to the geometric mean of the expected surprisal. As surprisal is itself in probability-space, this results in the surprisal that, if held constant across all timesteps, would yield the observed sequence surprisal, making it an effective length-normalized measure of total sequence surprisal.

The remaining BFR and 1TFR uncertainty measures are simple to define in terms of MC and FR measures. As described in further detail in Section \ref{sec:methods-inference}, our OEQA prompts are designed to encourage the model to output the selected answer immediately and mark the end of the chosen answer with a closing `\}' brace. The intention is to identify the most relevant answer segment so that further model explanation or otherwise over-generation can be excluded from UQ measurement. This is an especially important consideration for models that have not been instruct finetuned, as they often fail to generate a stop token, opting instead to generate additional questions until the maximum generation length is reached. BFR is exactly the base FR measures applied exclusively to the sequence segment that the model generated before the closing brace. We also hypothesized that most of the alignment signal may be further concentrated in the first generated token. We test this by measuring 1TFR measures, which is exactly the non-cloze MC measures applied to the first token generated by the model during OEQA. All three FR measure sets are calculated on the same generated output without regeneration.

\subsection{Model Inference}
\label{sec:methods-inference}
MCQA and OEQA prompts were designed similarly. The models were presented with two ICL examples, structured appropriately for MCQA or OEQA as necessary. Prior to the ICL examples, both templates also included a short system prompt, either in the designated system prompt location for supporting models or at the beginning of the context window, with instructional content to encourage models to answer the question immediately, without explanation, and ending with a closing brace to facilitate the BFR measures. After presenting the question, and answer choices for MCQA, the model's answer was artificially started with appropriate formatting for chat-finetuned models such that the first generated token is expected to contribute directly towards the intended answer rather than superfluous formatting. While this nudging might limit the generalization of the results herein to more complex tasks, it has the added benefit of better isolating to the uncertainty behavior on specifically the tokens that contribute to the answer without noise from framing tokens.

\begin{figure*}[t]
        \centering
        \includegraphics[width=\linewidth]{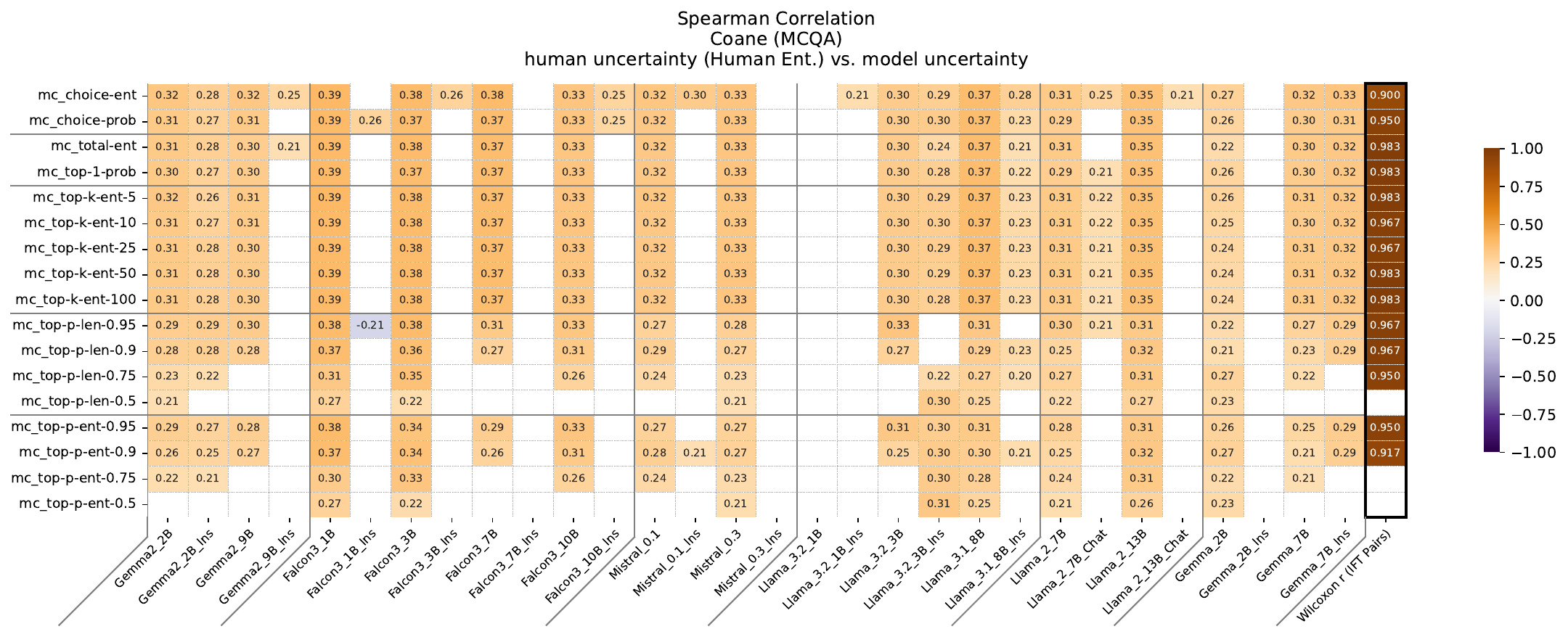}
        \caption{Spearman correlation between uncertainty measures and human uncertainty (measured via human response distribution entropy) on the MCQA version of the Coane dataset. Darker cells indicate higher correlation, while blank cells indicate that correlation failed to meet the corrected p-value threshold. The rightmost column shows the Wilcoxon effect size between each pair of base vs instruct models. Models are grouped by family, models are sorted within family by model size, and families are sorted by mean per-column maximum correlation}
        \label{fig:align-corr-coane-mcqa-hent}
    \end{figure*}

Model inference was performed using 5 A100 GPU hours in a container with identical methods across all MCQA datasets with minimal adaptation to accommodate OEQA for the Coane dataset. Where possible, each model's starting context was formatted in line with trained expectations as defined in the associated tokenizer chat template as provided by the huggingface apply\_chat\_template method. For models that support system prompts, the instructional content portion of the chat template was instead included in the system prompt rather than the main prompt body. Model-specific template preparations were also applied the ICL examples to maximize consistency.

All contexts were dynamically batched after formatting 
% and independently between models 
such that all contexts within a given batch were identical in terms of tokenized length. This ensured that no contexts included padding tokens which might have affected absolute token probabilities. For OEQA tasks, padding tokens may have been generated with batches yielded differing generation lengths, but padding tokens were stripped before UQ measure calculation.

Model correctness, necessary for downstream ECE calculation, required substantially different methods between the MCQA and OEQA tasks. MCQA correctness was readily obtainable with the model's chosen answer is taken to be the answer choice to which the highest total probability is assigned among the available answer choices. This chosen answer label is then compared against the expected answer label to determine correctness. To account for synonymous variations in label (e.g. ``A'' vs. ``A.'', ``A)'', ``a'', etc.), we sum across a collection of reasonable synonymous variants such that the probability of answer A is taken the be $P(\text{`}A\text{'})+P(\text{`}A.\text{'})+P(\text{`}A)\text{'})+P(\text{`}a.\text{'})+\dots$ and similarly for all MCQA answer labels.

% \begin{figure*}[t]
%         \centering
%         \includegraphics[width=\linewidth]{figures/agreement_heatmap.pdf}
%         \caption{Model agreement with correct answers (fact) or most common human answers (human) across all datasets. Brighter squares indicate a higher level of correctness/agreement percentage.}
%         \label{fig:agreement}
%     \end{figure*}

OEQA correctness was instead judged manually for correctness individually by two human annotators by comparing model generations against the expected answer, with disagreements between annotators being resolved by a third annotator. Reasonable allowances were made for spelling, grammar, synonymy, and correct alternatives for under-specific questions. During this process, two questions in the Coane dataset were identified as unsuitable for inclusion. Details and justification on these two are briefly presented in Appendix \ref{app:data-alts}. These two questions were excluded from all downstream analysis, with the exception of the model probing described in Section \ref{sec:methods-probing}, due to the late discovery. The small number of affected questions is unlikely to have substantially affected the final results, however.

    \begin{figure*}[h]
        \centering
        \includegraphics[width=\linewidth]{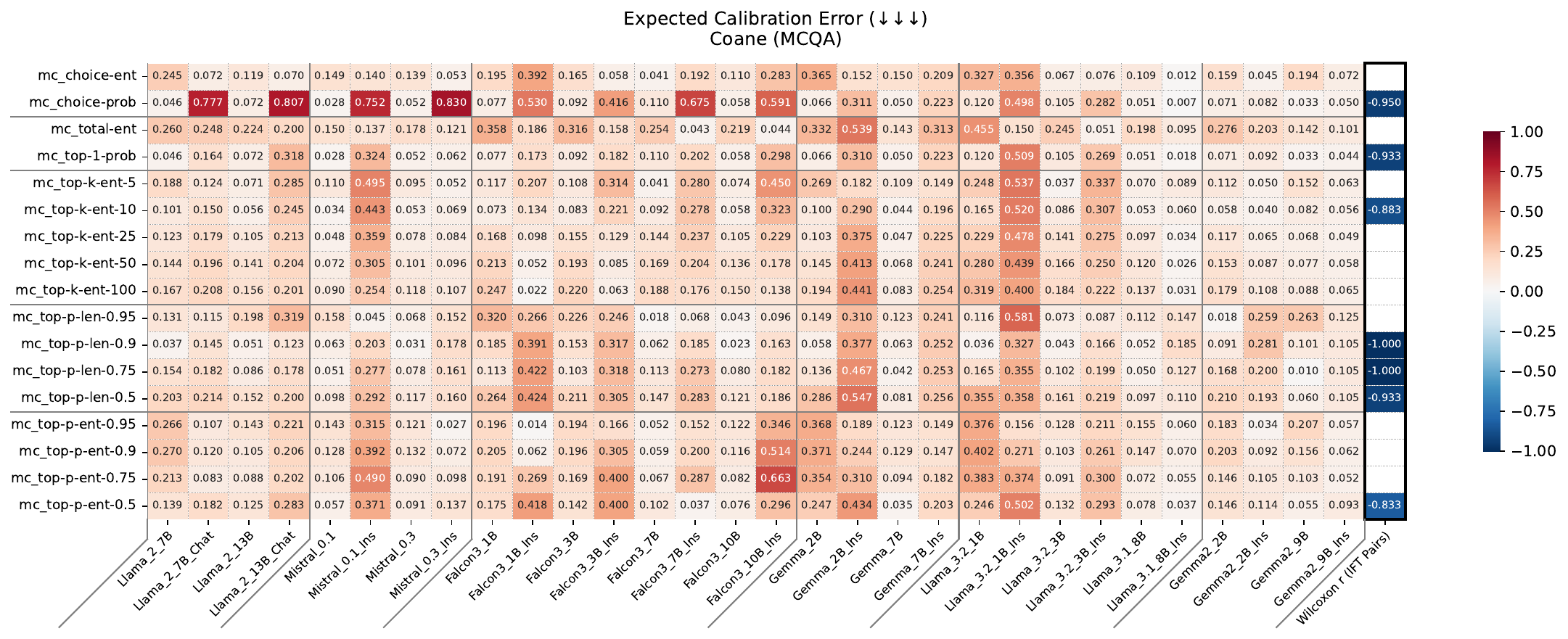}
        \caption{ECESweep results for the MCQA version of the Coane dataset. Darker cells indicate higher calibration error and thus lower calibration. The rightmost column shows the Wilcoxon effect size between each pair of base vs instruct models, with blank cells indicating insignificance. Models are grouped by family, models are sorted within family by model size, and families are sorted by mean per-column maximum correlation}
        \label{fig:ece-coane-mcqa}
    \end{figure*}
    
\iffalse
* Shared:
    
* MCQA-specific:
    * cloze test for determining model answer
    * pooled probabilities of plausible alternative formats (e.g. "A" vs "a" vs "A)", etc.)
* FR-specific:
    * model answers manually judged for correctness by 2 human annotators, with a third human annotator to resolve disagreement

will likely be worth making explicit that the prompt nudges (ICL, model response start spoofing, etc.) could impact the generalization of the results to less structured tasks and contexts. In our defense, however, this is an effective way to isolate to the actual behavior we are interested in. Specifically, we isolate the task to the task the human would be attempting to solve (factual question answering) by artificially bypassing secondary tasks like rote instruction following. This ensures that we are measuring uncertainty during performance of the as close to the same task as the human comparison group.
\fi

\subsection{Model Probing}
\label{sec:methods-probing}
    To investigate where and how uncertainty alignment presents in the LLM internals, we use a probing method inspired by \citet{burger2024truth}, in which we train linear regression models that attempt to predict human uncertainty from LLM activation at each layer. For a given LLM, we presented the Coane MCQA formatted contexts to the LLM and extracted the last-token activations for each context. We used the resulting activations, to train a simple linear regression model for each layer of the LLM with human uncertainty as the regression target. A separate model was trained for each of three measures of human MCQA uncertainty: response distribution entropy, average response time when correct, and average overall response time. We employed 10-fold cross validation when training each layer's linear regressors and evaluated the regressors using Pearson r correlation.

\section{Results}
In this section we present the results of our various analyses.

\subsection{Agreement and Correctness}
As an initial surface-level analysis, we calculated for each model, the model's correctness on factual datasets (Coane, CamChoice, MMLU) and the model's agreement with human answer preference for datasets with human response distribution data (Coane, CamChoice, ProtoQA). Model correctness on factual tasks scaled positively with parameter count in base and instruct models (r = 0.45–0.71 across factual datasets), with instruct-finetuned variants showing no systematic advantage over their base counterparts except in open-ended factual recall, where finetuning was associated with a significant decrease in accuracy (Wilcoxon p = 0.017, r = 0.67). Agreement with the most common human response followed a similar but attenuated pattern on knowledge-based datasets, and was largely independent of model size on the purely preference-based ProtoQA dataset

% The results are shown in Figure \ref{fig:agreement}. Model size seems to be the biggest predictor of factual accuracy, but also weakly tied to human agreement. To quantify this, Pearson correlations between parameter count and factual accuracy across base models range from moderate to strong (r = 0.45–0.71 across factual datasets), while Wilcoxon signed-rank tests comparing each base model to its instruct-finetuned counterpart reveal no significant systematic shift in accuracy for most conditions. The one exception is open-ended factual recall, where instruct finetuning is associated with a significant and large-magnitude decrease in accuracy (p = 0.017, r = 0.67). 

Given that no such size-dependence is discernable for the purely preference-base ProtoQA, however, this is likely only an artifact of size-dependent correctness and human accuracy. Generally, agreement per model is ranked such that correctness on MCQA tasks is higher than than both OEQA correctness and MCQA human-similarity. OEQA correctness tends higher than MCQA human-similarity for all nearly all models larger than 3B.

    \begin{figure*}[t]
    \centering
    \includegraphics[width=\linewidth]{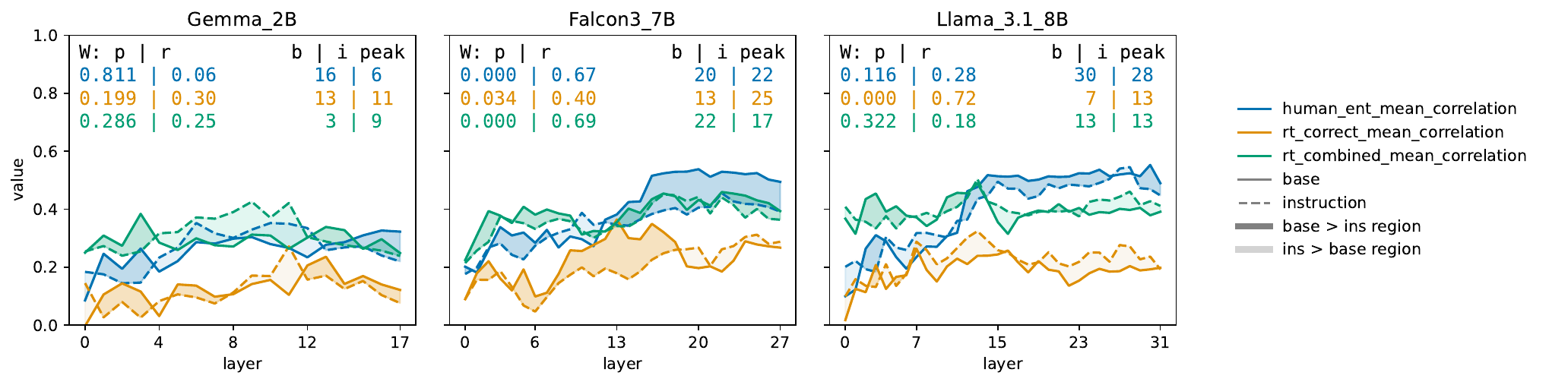}
    \caption{Per-layer model activation probing correlations. Colors indicate the human uncertainty type targeted during linear regression. Shaded regions indicate the gap between the base mode (solid line) and the instruct finetuned variant (dashed line). On-figure annotations in the top-left indicate the p-value and effect size for a Wilcoxon rank sum test between the base correlation and ins correlation across layers. Annotations in the top-right display the layer at which maximum human alignment was measured.}
    \label{fig:internal_subset}
\end{figure*}

\subsection{Uncertainty Alignment}
\label{sec:res-align}

    The rightmost column of Figure \ref{fig:align-corr-coane-mcqa-hent} displays the significant effect sizes per measure for Wilcoxon signed rank tests between the alignment of each base model versus the alignment of its instruct-finetuned counterpart. Across nearly every measure, this test finds a strong relationship such that instruct finetuned models tend to be less aligned that their base counterparts.
    
    The remaining results for other all other dataset, task, and human uncertainty configurations can be found in Figures \ref{fig:app-corr-first}-\ref{fig:app-corr-last} in Appendix \ref{app:addt-figs}. In brief, the noteworthy trends are that the alignment signal is present but far more sporadically and model-dependent for the CamChoice dataset and for MCQA response time correlation on Coane. ProtoQA shows consistently significant, but very weak alignment. For OEQA, the models show nearly no significant evidence of alignment for any of the FR uncertainty measure classes. Only Falcon 3 10B shows any significant correlation with human response time, but this may be an artifact of extraneous generation behavior as the same model shows no alignment for BFR or 1TFR.

    Unlike factual accuracy, uncertainty alignment shows no consistent relationship with model size in either base or instruct models. 
    
\subsection{Uncertainty Calibration}

    Following up on its standout alignment in the prior section, we have evaluated the ECE scores for each MCQA measure-model pair on the Coane dataset in Figure \ref{fig:ece-coane-mcqa}, with the remaining datasets and Coane OEQA measures presented in Figures \ref{fig:app-ece-first}-\ref{fig:app-ece-last} in Appendix \ref{app:addt-figs}. With a few noteworthy exceptions, the results show relatively strong calibration on the Coane dataset, with every model showing strong calibration (ECE $\lesssim 0.1$) for at least one measure. Like with alignment, there is noticeable model-dependency, with the Mistral 0.1 Ins, Gemma 2B Ins, and LLaMa 3.2 1B Ins showing substantially worse calibration across nearly every measure than most other models. For measures, mc\_choice-prob stands out as particularly volatile, being highly calibrated for most models, but actively anti-calibrated for the instruct-finetuned variants of all LLaMa 2 and Mistral models. The Wilcoxon column shows evidence of significant degradation in calibration after instruct finetuning across many measures. This result, while initially surprising, is consistent with existing findings on the unintended effects of instruct finetuning \citep{huang2025calibrated}. Taken together with the results of section \ref{sec:res-align}, this suggests that alignment and calibration are both negatively affected by instruct finetuning.

    Unlike alignment, calibration appears less affected by choice of model and dataset, with lower but still substantial calibration across all pairings for both MMLU and CamChoice. This may be explainable by the relative difficulty of these two datasets when compared with Coane MCQA. Unlike the alignment results, the OEQA results on Coane show strong calibration across all three uncertainty measure groups. Average top-k entropy in particular shows strong calibration for full generation measures with relatively little loss in BFR. In a noteworthy deviation from the MCQA case, we observe a significant positive Wilcoxon effect size for some measures, indicating some measures may improve calibration with instruct finetuning outside of limited-option contexts.

    As a final point of comparison, we calculated the human calibration on the Coane dataset. Because we have access to per-question accuracies, we are able to sidestep binning questions by taking average instance-level calibration $\frac{1}{|Q|}\sum_{q\in Q}|Acc(q)-Conf(q)|$ where $Q$ is the set of all questions. The results were average calibration errors of $0.226$ for MCQA with human entropy, $0.210$ for MCQA with response time, and $0.445$ for OEQA with response time. This indicates that the LLMs tend to be more well-calibrated than human subjects across most UQ measures.
    
\subsection{Probing}

The results of probing the LLM internals are shown in Figure \ref{fig:internal_subset} for a representative subset of models, with the remaining models reported in Figure \ref{fig:app-probing} in Appendix \ref{app:addt-figs}. In addition to plotting the correlation across layers, we perform a Wilcoxon signed-rank test on the per layer base model correlation vs instruct model correlations. We also identify the layers at which maximum correlation is attained with each of the human uncertainty measures.

Consistently across nearly every model, the most correlated human measure with model activations, both in absolute terms over all layers and in terms of final layer activations, is human entropy. This dominance tends to arise near the middle layers. Notably, this occurs because of the relatively flat trajectory of the response time across layers, while human entropy tends to become more correlated with depth. This suggests a meaningful difference in how the models internalize human individual-level (RT) vs group-level (entropy) uncertainty. Group uncertainty alignment consistently occurs stronger, later, and with increasing strength throughout inference. We also see, when comparing against the alignment heatmaps in Section \ref{sec:res-align}, that maximum correlation for every model tends to be substantially higher than that observed only in the output logits. For an extreme example, Gemma 2 9B shows a maximum of $r\approx0.6$ correlation with human entropy when using the internal probe, but a maximum of $r=0.32$ correlation for logit-based measures. This shows promise for further behavioral identification and isolation in future LLM behavioral studies.

\section{Conclusion}
In this work, we have extensively studied the uncertainty behavior of an array of pre-trained LLMs. We have verified and extended previous results by finding evidence of a consistent, but moderate in strength and sensitive to measurement methodology, positive uncertainty alignment signal. We have demonstrated that models are often simultaneously aligned and calibrated in their uncertainty and that both of these tend to be disrupted by instruction finetuning. We also showed through activation probing that LLM overt behavior does not fully capture the model alignment and found initial evidence that human similarity may be better characterized in terms of similarity to groups of humans rather than representative individuals.

\section{Future Work}
As the field of uncertainty alignment in LLMs remains largely unexplored, there are many viable paths for further exploration in future work. This work has remained limited to the narrow tasks of short-form question answering. Future work should expand into more complex tasks, such as problem-solving, planning, or long-form response tasks. To bring our work more in line with existing calibration research, future work should also investigate multi-inference and other more complex methods of uncertainty quantification. Finally, with the recognition of uncertainty alignment signals within model activation patterns, isolation of the related latent vectors may allow for direct manipulation, enhancement, or even reversal of alignment behaviors without significant retraining, broadening to possible applications of the research presented herein.

\section*{Limitations}
The largest limitation of the study herein is the relatively limited task set, covering only simple factual recall tasks. It is difficult to guarantee generalization of any results reported herein to more complex tasks, especially in light of the nearly non-existent overt alignment results on the OEQA task. Our most well-performing dataset, Coane, is also targeted at an age-specific cohort of older adults. This may harm generalization to more diverse human comparison sets, though it is also possible that the knowledge biases within this group may have artificially deflated the final results. 

% \section*{Ethical Considerations}
% Aside from the standard concerns regarding power and resource usage during model inference, the largest ethical consideration surrounding this work is the implications human aligned uncertainty may have on the ability to manipulate or mislead human users and the danger of economic replacement by uncertainty aligned AI. For the former, a model that is more convincing and rapport-building in it's uncertainty reporting without consummate improvements to calibration may convincingly mislead users on their actual efficacy. For the latter, we hope that our work contributes to greater human-AI collaboration, as opposed to replacement, but human similarity in task coverage with above-human task calibration and efficacy may be a harmful combination for at-risk workers.

% \section*{Acknowledgments}
% Optional Section. Not clear whether it contributes to page limit


\bibliography{custom}

@article{hick1952rate,
  title={On the rate of gain of information},
  author={Hick, William E},
  journal={Quarterly Journal of experimental psychology},
  volume={4},
  number={1},
  pages={11--26},
  year={1952},
  publisher={Taylor \& Francis}
}

@article{Moore_Roberts_Watson_Wisniewski_2026, title={Investigating Human-Aligned Large Language Model Uncertainty}, volume={39}, url={https://journals.flvc.org/FLAIRS/article/view/141835}, DOI={10.32473/flairs.39.1.141835}, abstractNote={&amp;lt;p&amp;gt;Recent work has sought to quantify large language model uncertainty to facilitate model control and modulate user trust. Previous works focus on measures of uncertainty that are theoretically grounded or reflect the average overt behavior of the model. In this work, we investigate a variety of uncertainty measures, in order to identify measures that correlate with human group-level uncertainty. We find that Bayesian measures and a variation on entropy measures, top k entropy, tend to agree with human behavior as a function of model size. We find that some strong measures decrease in human-similarity with model size, but, by multiple linear regression, we find that combining multiple uncertainty measures provide comparable human-alignment with reduced size-dependency.&amp;lt;/p&amp;gt;}, number={1}, journal={The International FLAIRS Conference Proceedings}, author={Moore, Kyle and Roberts, Jesse and Watson, Daryl and Wisniewski, Pamela}, year={2026}, month={May} }

@misc{Falcon3,
    title  = {The Falcon 3 Family of Open Models},
    url    = {https://huggingface.co/blog/falcon3},
    author = {Falcon-LLM Team},
    month  = {December},
    year   = {2024}
}

@inproceedings{liu2025uncertainty,
  title={Uncertainty quantification and confidence calibration in large language models: A survey},
  author={Liu, Xiaoou and Chen, Tiejin and Da, Longchao and Chen, Chacha and Lin, Zhen and Wei, Hua},
  booktitle={Proceedings of the 31st ACM SIGKDD Conference on Knowledge Discovery and Data Mining V. 2},
  pages={6107--6117},
  year={2025}
}

@article{sun2025large,
  title={Large language models are overconfident and amplify human bias},
  author={Sun, Fengfei and Li, Ningke and Wang, Kailong and Goette, Lorenz},
  journal={arXiv preprint arXiv:2505.02151},
  year={2025}
}

@article{steyvers2025large,
  title={What large language models know and what people think they know},
  author={Steyvers, Mark and Tejeda, Heliodoro and Kumar, Aakriti and Belem, Catarina and Karny, Sheer and Hu, Xinyue and Mayer, Lukas W and Smyth, Padhraic},
  journal={Nature Machine Intelligence},
  volume={7},
  number={2},
  pages={221--231},
  year={2025},
  publisher={Nature Publishing Group UK London}
}

@inproceedings{roberts2024large,
  title={Large language model recall uncertainty is modulated by the fan effect},
  author={Roberts, Jesse and Moore, Kyle and Fisher, Douglas and Ewaleifoh, Oseremhen and Pham, Thao},
  booktitle={Proceedings of the 28th conference on computational natural language learning},
  pages={303--313},
  year={2024}
}

@inproceedings{moore2025chain,
  title={Chain of Thought Still Thinks Fast: APriCoT Helps with Thinking Slow},
  author={Moore, Kyle and Roberts, Jesse and Pham, Thao Thi Minh and Fisher, Douglas},
  booktitle={Proceedings of the Annual Meeting of the Cognitive Science Society},
  volume={47},
  year={2025}
}

@inproceedings{xu2025language,
  title={Do language models mirror human confidence? exploring psychological insights to address overconfidence in LLMs},
  author={Xu, Chenjun and Wen, Bingbing and Han, Bin and Wolfe, Robert and Wang, Lucy Lu and Howe, Bill},
  booktitle={Findings of the Association for Computational Linguistics: ACL 2025},
  pages={25655--25672},
  year={2025}
}

@inproceedings{guo2017calibration,
  title={On calibration of modern neural networks},
  author={Guo, Chuan and Pleiss, Geoff and Sun, Yu and Weinberger, Kilian Q},
  booktitle={International conference on machine learning},
  pages={1321--1330},
  year={2017},
  organization={PMLR}
}

@article{wang2023calibration,
  title={Calibration in deep learning: A survey of the state-of-the-art},
  author={Wang, Cheng},
  journal={arXiv preprint arXiv:2308.01222},
  year={2023}
}

@inproceedings{roelofs2022mitigating,
  title={Mitigating bias in calibration error estimation},
  author={Roelofs, Rebecca and Cain, Nicholas and Shlens, Jonathon and Mozer, Michael C},
  booktitle={International Conference on Artificial Intelligence and Statistics},
  pages={4036--4054},
  year={2022},
  organization={PMLR}
}

@article{touvron2023llama,
  title={Llama 2: Open foundation and fine-tuned chat models},
  author={Touvron, Hugo and Martin, Louis and Stone, Kevin and Albert, Peter and Almahairi, Amjad and Babaei, Yasmine and Bashlykov, Nikolay and Batra, Soumya and Bhargava, Prajjwal and Bhosale, Shruti and others},
  journal={arXiv preprint arXiv:2307.09288},
  year={2023}
}

@article{grattafiori2024llama,
  title={The llama 3 herd of models},
  author={Grattafiori, Aaron and Dubey, Abhimanyu and Jauhri, Abhinav and Pandey, Abhinav and Kadian, Abhishek and Al-Dahle, Ahmad and Letman, Aiesha and Mathur, Akhil and Schelten, Alan and Vaughan, Alex and others},
  journal={arXiv preprint arXiv:2407.21783},
  year={2024}
}

@misc{jiang2023mistral7b,
      title={Mistral 7B}, 
      author={Albert Q. Jiang and Alexandre Sablayrolles and Arthur Mensch and Chris Bamford and Devendra Singh Chaplot and Diego de las Casas and Florian Bressand and Gianna Lengyel and Guillaume Lample and Lucile Saulnier and Lélio Renard Lavaud and Marie-Anne Lachaux and Pierre Stock and Teven Le Scao and Thibaut Lavril and Thomas Wang and Timothée Lacroix and William El Sayed},
      year={2023},
      eprint={2310.06825},
      archivePrefix={arXiv},
      primaryClass={cs.CL},
      url={https://arxiv.org/abs/2310.06825}, 
}

@article{team2024gemma,
  title={Gemma: Open models based on gemini research and technology},
  author={Team, Gemma and Mesnard, Thomas and Hardin, Cassidy and Dadashi, Robert and Bhupatiraju, Surya and Pathak, Shreya and Sifre, Laurent and Rivi{\`e}re, Morgane and Kale, Mihir Sanjay and Love, Juliette and others},
  journal={arXiv preprint arXiv:2403.08295},
  year={2024}
}

@article{team2024gemma2,
  title={Gemma 2: Improving open language models at a practical size},
  author={Team, Gemma and Riviere, Morgane and Pathak, Shreya and Sessa, Pier Giuseppe and Hardin, Cassidy and Bhupatiraju, Surya and Hussenot, L{\'e}onard and Mesnard, Thomas and Shahriari, Bobak and Ram{\'e}, Alexandre and others},
  journal={arXiv preprint arXiv:2408.00118},
  year={2024}
}

@article{hendrycks2020measuring,
  title={Measuring massive multitask language understanding},
  author={Hendrycks, Dan and Burns, Collin and Basart, Steven and Zou, Andy and Mazeika, Mantas and Song, Dawn and Steinhardt, Jacob},
  journal={arXiv preprint arXiv:2009.03300},
  year={2020}
}

@inproceedings{boratko2020protoqa,
  title={ProtoQA: A question answering dataset for prototypical common-sense reasoning},
  author={Boratko, Michael and Li, Xiang and O’Gorman, Tim and Das, Rajarshi and Le, Dan and McCallum, Andrew},
  booktitle={Proceedings of the 2020 Conference on Empirical Methods in Natural Language Processing (EMNLP)},
  pages={1122--1136},
  year={2020}
}

@article{farquhar2024detecting,
  title={Detecting hallucinations in large language models using semantic entropy},
  author={Farquhar, Sebastian and Kossen, Jannik and Kuhn, Lorenz and Gal, Yarin},
  journal={Nature},
  volume={630},
  number={8017},
  pages={625--630},
  year={2024},
  publisher={Nature Publishing Group UK London}
}

@article{shorinwa2025survey,
  title={A survey on uncertainty quantification of large language models: Taxonomy, open research challenges, and future directions},
  author={Shorinwa, Ola and Mei, Zhiting and Lidard, Justin and Ren, Allen Z and Majumdar, Anirudha},
  journal={ACM Computing Surveys},
  volume={58},
  number={3},
  pages={1--38},
  year={2025},
  publisher={ACM New York, NY}
}

@article{huang2025calibrated,
  title={Calibrated language models and how to find them with label smoothing},
  author={Huang, Jerry and Lu, Peng and Zeng, Qiuhao},
  journal={arXiv preprint arXiv:2508.00264},
  year={2025}
}

@article{coane2021database,
  title={A database of general knowledge question performance in older adults},
  author={Coane, Jennifer H and Umanath, Sharda},
  journal={Behavior Research Methods},
  volume={53},
  number={1},
  pages={415--429},
  year={2021},
  publisher={Springer}
}

@article{mullooly2023cambridge,
  title={The cambridge multiple-choice questions reading dataset},
  author={Mullooly, Andrew and Andersen, {\O}istein and Benedetto, Luca and Buttery, Paula and Caines, Andrew and Gales, Mark JF and Karatay, Yasin and Knill, Kate and Liusie, Adian and Raina, Vatsal and others},
  year={2023},
  publisher={Cambridge University Press and Assessment}
}

@article{huang2025look,
  title={Look before you leap: An exploratory study of uncertainty analysis for large language models},
  author={Huang, Yuheng and Song, Jiayang and Wang, Zhijie and Zhao, Shengming and Chen, Huaming and Juefei-Xu, Felix and Ma, Lei},
  journal={IEEE Transactions on Software Engineering},
  volume={51},
  number={2},
  pages={413--429},
  year={2025},
  publisher={IEEE}
}

@article{burger2024truth,
  title={Truth is universal: Robust detection of lies in llms},
  author={B{\"u}rger, Lennart and Hamprecht, Fred A and Nadler, Boaz},
  journal={Advances in Neural Information Processing Systems},
  volume={37},
  pages={138393--138431},
  year={2024}
}

\newpage
\appendix

\section{Dataset Alterations}
\label{app:data-alts}
In this section, we briefly detail minor alterations and/or exclusions made to the datasets. The majority of changes were small encoding changes, specifically the replacement of some Unicode punctuation characters with their ASCII equivalents. This was done to guard against interference in model behavior from the inclusion of equivalent but much rarer tokens.

Most notably, two questions were removed from the Coane dataset. The first of these concerned the current holder of a public political office. This was removed for two-fold reasons. First, the correct answer to this question has changed since the original human trials, meaning that the correct answer as marked is outdated. While this could have been rectified, we removed the question entirely because the wide range of knowledge cutoffs across models makes it very likely that the ``correct'' accessible answer varies widely. The second question concerned a detail to a popular fantasy book, ``The Little Prince''. On further research, the annotators concluded that not only is the answer provided in the original dataset not correct based on publicly accessible information on the story, but the question asked has no non-trivial correct alternative answer. As such, it was also removed from all further analysis.

\section{Model Prompt Examples}
This section will contain example prompts, including question and answer choices where appropriate, as they were presented to the models. The specific formats presented are those used for the Falcon 3 1B base model. In all cases, underscores (\_) indicate the model's first token generation position and are not included in the context provided to the mode.

\subsection{ProtoQA}
\begin{verbatim}
Question: What is the chemical symbol for
gold?
A. Ag
B. Mo
C. Au
D. Gd
Answer: C

Question: In what year did the First World
War end?
A. 1939
B. 1918
C. 1914
D. 1945
Answer: B

Question: At The Beach, Name Something
That Might Protect You From Sun.
A. umbrella
B. sunscreen
C. sun hat
D. sunglasses
E. cover up
F. shade
Answer: _
\end{verbatim}

\subsection{CamChoice}
\begin{verbatim}
    
Question: What is the chemical symbol for
gold?
A. Ag
B. Mo
C. Au
D. Gd
Answer: C

Question: In what year did the First World
War end?
A. 1939
B. 1918
C. 1914
D. 1945
Answer: B

Question: Some time ago a website
highlighted the risks of public check-ins
- online announcements of your whereabouts.
...
<story truncated for example purposes only>
...
What Armstrong has figured out is something
we would all do well to remember: the web
may allow us to say anything, but that
doesn't mean we should. Why does the writer
describe a website about public check-ins
in the first paragraph?
A. to reinforce the concerns already felt
by some people
B. to remind readers to beware of false
promises
C. to explain that such sites often have a
hidden agenda
D. to show that the risks of internet use
are sometimes over estimated
Answer: _
\end{verbatim}

\subsection{MMLU}
\begin{verbatim}
Question: What is the chemical symbol for
gold?
A. Ag
B. Mo
C. Au
D. Gd
Answer: C

Question: In what year did the First
World War end?
A. 1939
B. 1918
C. 1914
D. 1945
Answer: B

Question: Find the degree for the given
field extension Q(sqrt(2), sqrt(3),
sqrt(18)) over Q.
A. 0
B. 4
C. 2
D. 6
Answer: _
\end{verbatim}

\subsection{Coane}
\subsubsection{MCQA}
\begin{verbatim}
Question: What is the chemical symbol for
gold?
A. Ag
B. Mo
C. Au
D. Gd
Answer: C

Question: In what year did the First World
War end?
A. 1939
B. 1918
C. 1914
D. 1945
Answer: B

Question: Which band was Paul McCartney a
member of?
A. The Monkees
B. The Rolling Stones
C. The Yardbirds
D. The Beatles
Answer: _
\end{verbatim}

\subsubsection{OEQA}
\begin{verbatim}
What is the chemical symbol for
gold?
Answer in as few words as possible,
wrapping the key part in {} brackets:
The chemical symbol for gold is {Au}.

In what year did the First World
War end?
Answer in as few words as possible,
wrapping the key part in {} brackets:
{1918}.

Which band was Paul McCartney a
member of?
Answer in as few words as possible,
wrapping the key part in {}
brackets: {_
\end{verbatim}

\section{Additional Figures}
\label{app:addt-figs}

\begin{figure*}
    \centering
    \includegraphics[width=\linewidth]{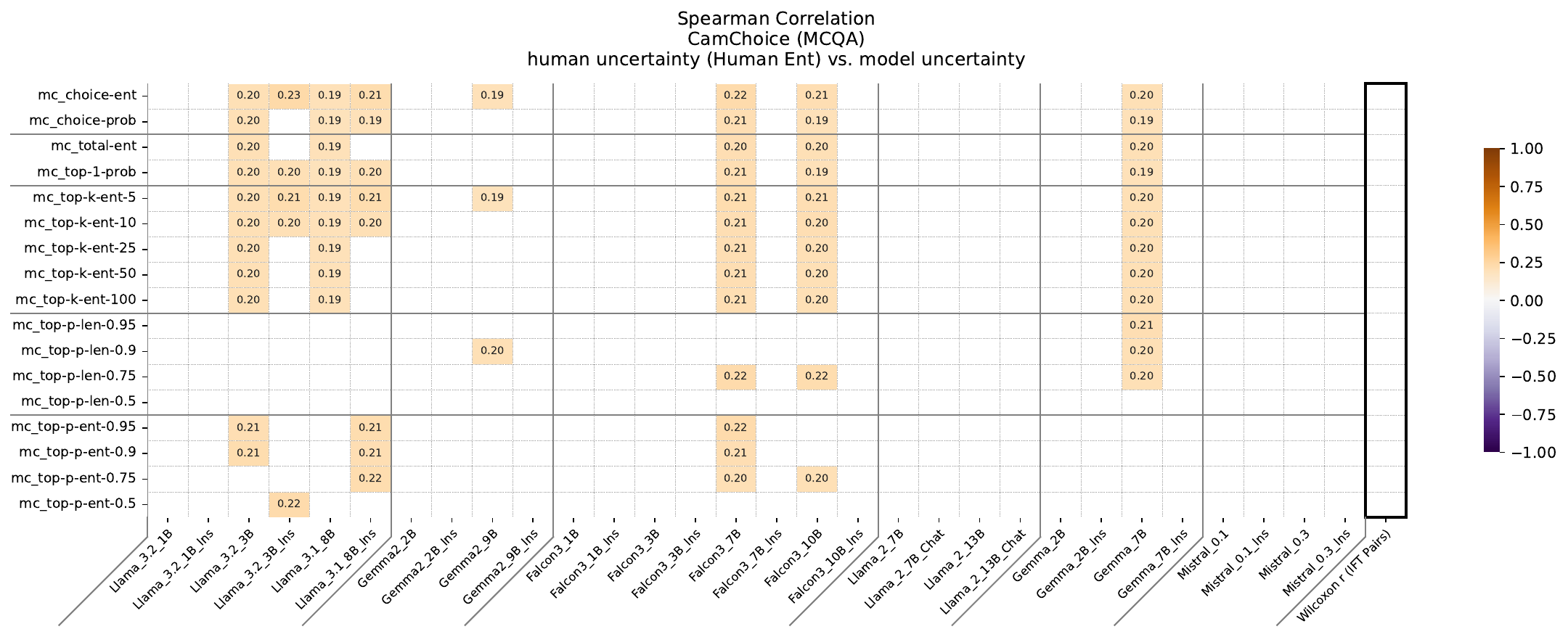}
    \caption{Spearman correlation between uncertainty measures and human uncertainty (measured via human response distribution entropy) on the CamChoice dataset. Darker cells indicate higher correlation, while blank cells indicate that correlation failed to meet the corrected p-value threshold. The rightmost column shows the Wilcoxon effect size between each pair of base vs instruct models. Models are grouped by family, models are sorted within family by model size, and families are sorted by mean per-column maximum correlation}
    \label{fig:app-corr-first}
\end{figure*}
\begin{figure*}
    \centering
    \includegraphics[width=\linewidth]{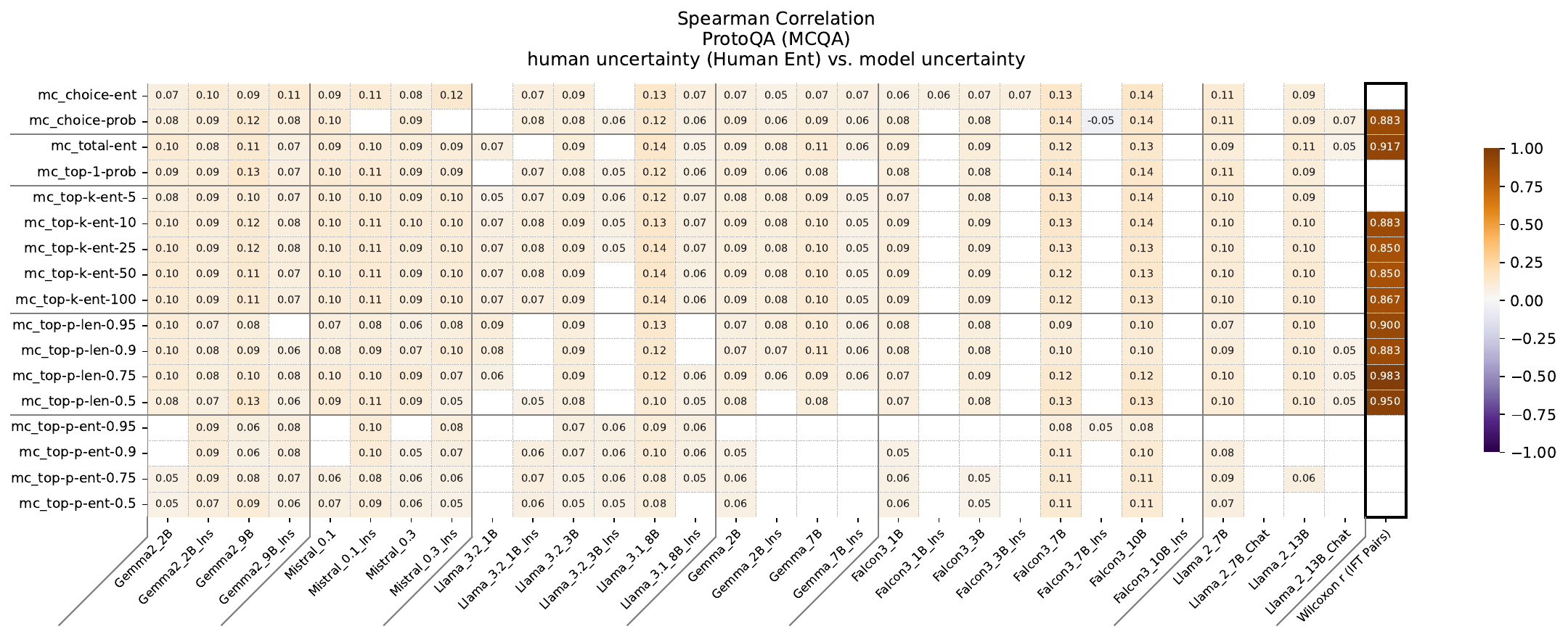}
    \caption{Spearman correlation between uncertainty measures and human uncertainty (measured via human response distribution entropy) on the ProtoQA dataset. Darker cells indicate higher correlation, while blank cells indicate that correlation failed to meet the corrected p-value threshold. The rightmost column shows the Wilcoxon effect size between each pair of base vs instruct models. Models are grouped by family, models are sorted within family by model size, and families are sorted by mean per-column maximum correlation}
    \label{fig:app-corr-protoqa}
\end{figure*}
\begin{figure*}
    \centering
    \includegraphics[width=\linewidth]{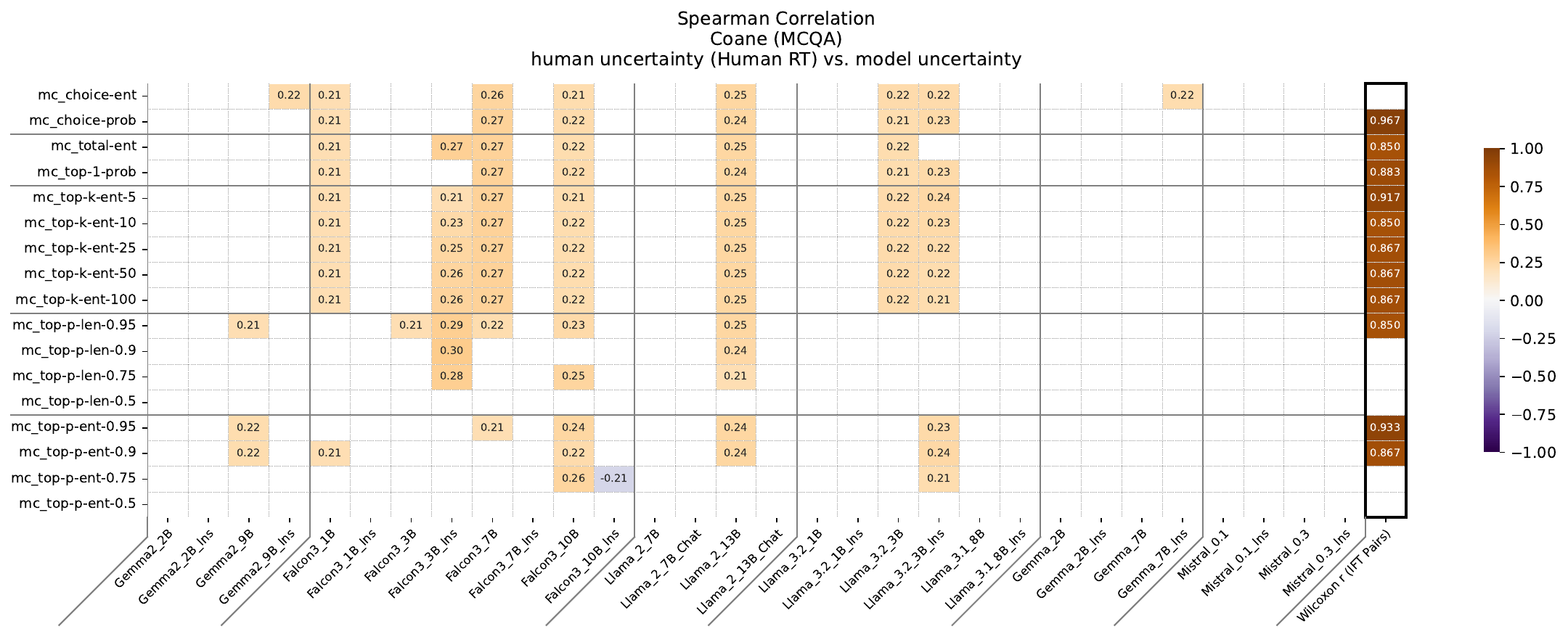}
    \caption{Spearman correlation between uncertainty measures and human uncertainty (measured via human response distribution entropy) on the MCQA version of the Coane dataset compared against human response time. Darker cells indicate higher correlation, while blank cells indicate that correlation failed to meet the corrected p-value threshold. The rightmost column shows the Wilcoxon effect size between each pair of base vs instruct models. Models are grouped by family, models are sorted within family by model size, and families are sorted by mean per-column maximum correlation}
\end{figure*}
\begin{figure*}
    \centering
    \includegraphics[width=\linewidth]{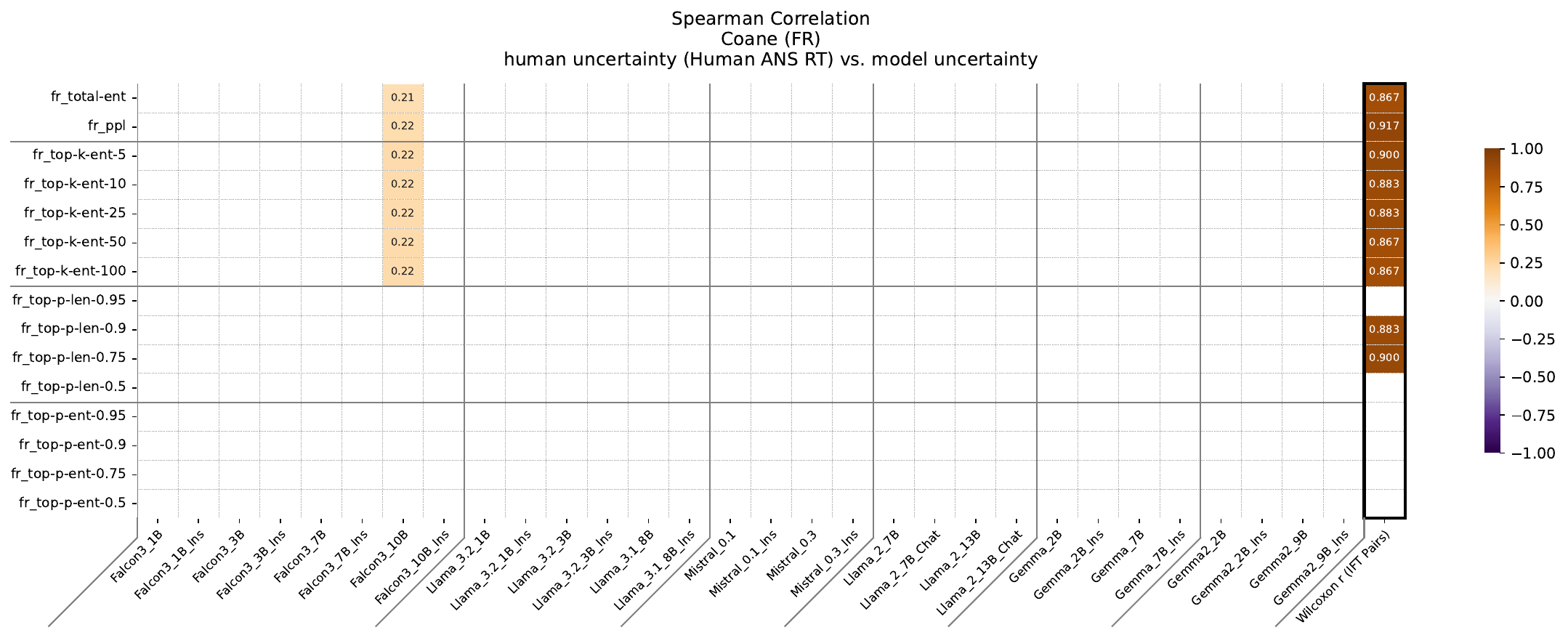}
    \caption{Spearman correlation between uncertainty measures and human uncertainty (measured via human response distribution entropy) on the FR version of the Coane dataset. Darker cells indicate higher correlation, while blank cells indicate that correlation failed to meet the corrected p-value threshold. The rightmost column shows the Wilcoxon effect size between each pair of base vs instruct models. Models are grouped by family, models are sorted within family by model size, and families are sorted by mean per-column maximum correlation}
\end{figure*}
\begin{figure*}
    \centering
    \includegraphics[width=\linewidth]{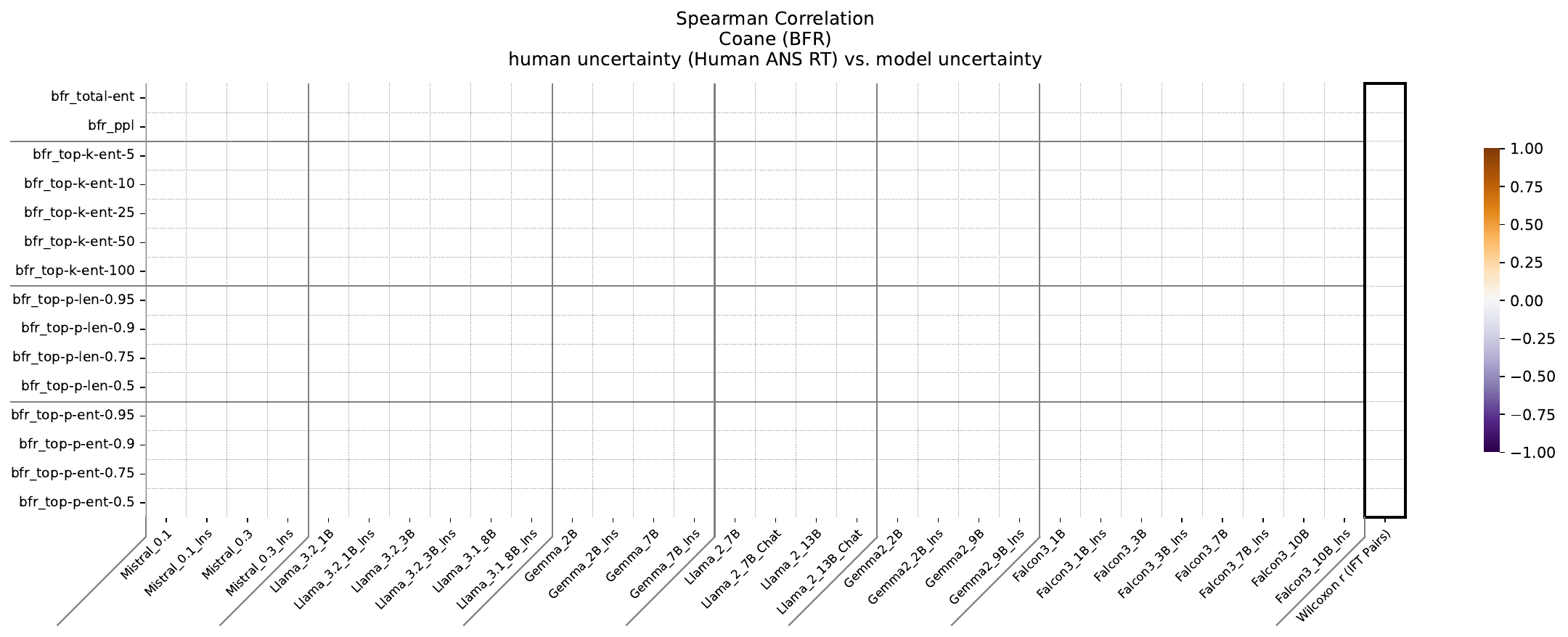}
    \caption{Spearman correlation between uncertainty measures and human uncertainty (measured via human response distribution entropy) on the BFR version of the Coane dataset. Darker cells indicate higher correlation, while blank cells indicate that correlation failed to meet the corrected p-value threshold. The rightmost column shows the Wilcoxon effect size between each pair of base vs instruct models. Models are grouped by family, models are sorted within family by model size, and families are sorted by mean per-column maximum correlation}
\end{figure*}
\begin{figure*}
    \centering
    \includegraphics[width=\linewidth]{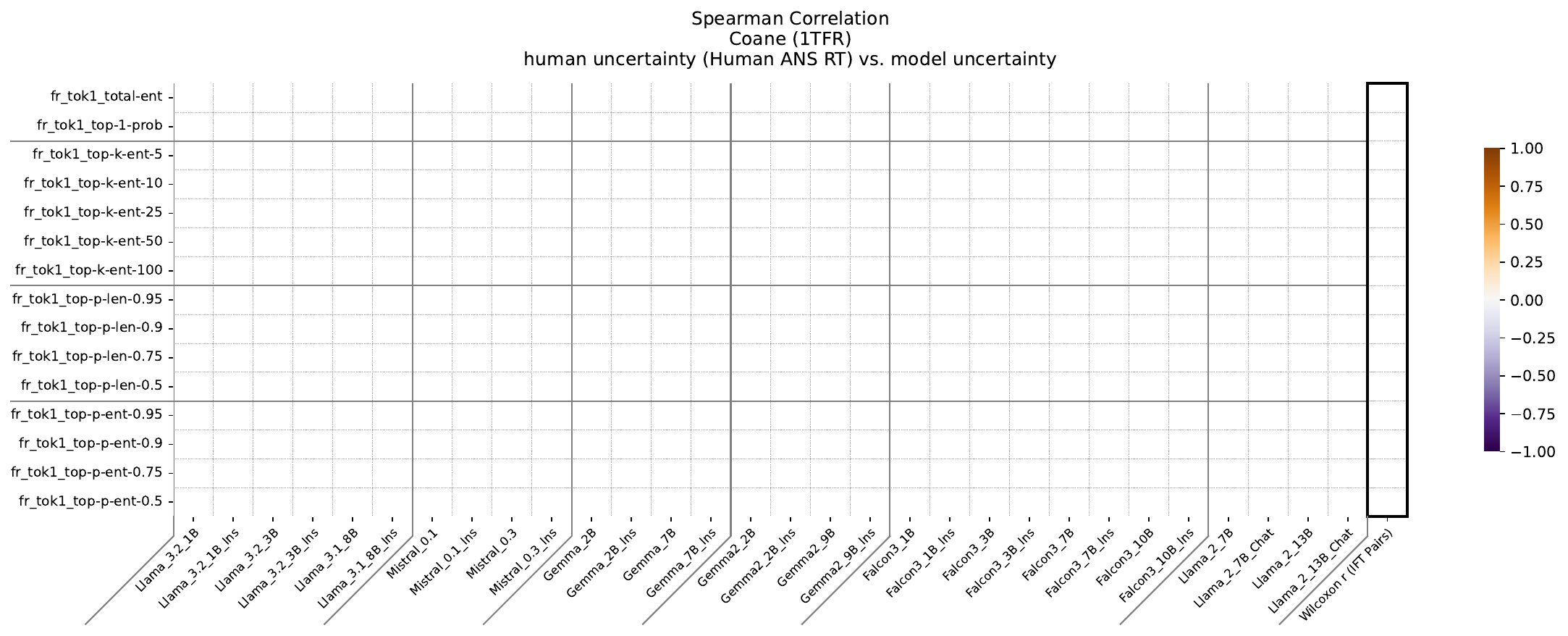}
    \caption{Spearman correlation between uncertainty measures and human uncertainty (measured via human response distribution entropy) on the 1TFR version of the Coane dataset. Darker cells indicate higher correlation, while blank cells indicate that correlation failed to meet the corrected p-value threshold. The rightmost column shows the Wilcoxon effect size between each pair of base vs instruct models. Models are grouped by family, models are sorted within family by model size, and families are sorted by mean per-column maximum correlation}
    \label{fig:app-corr-last}
\end{figure*}

\begin{figure*}
    \centering
    \includegraphics[width=\linewidth]{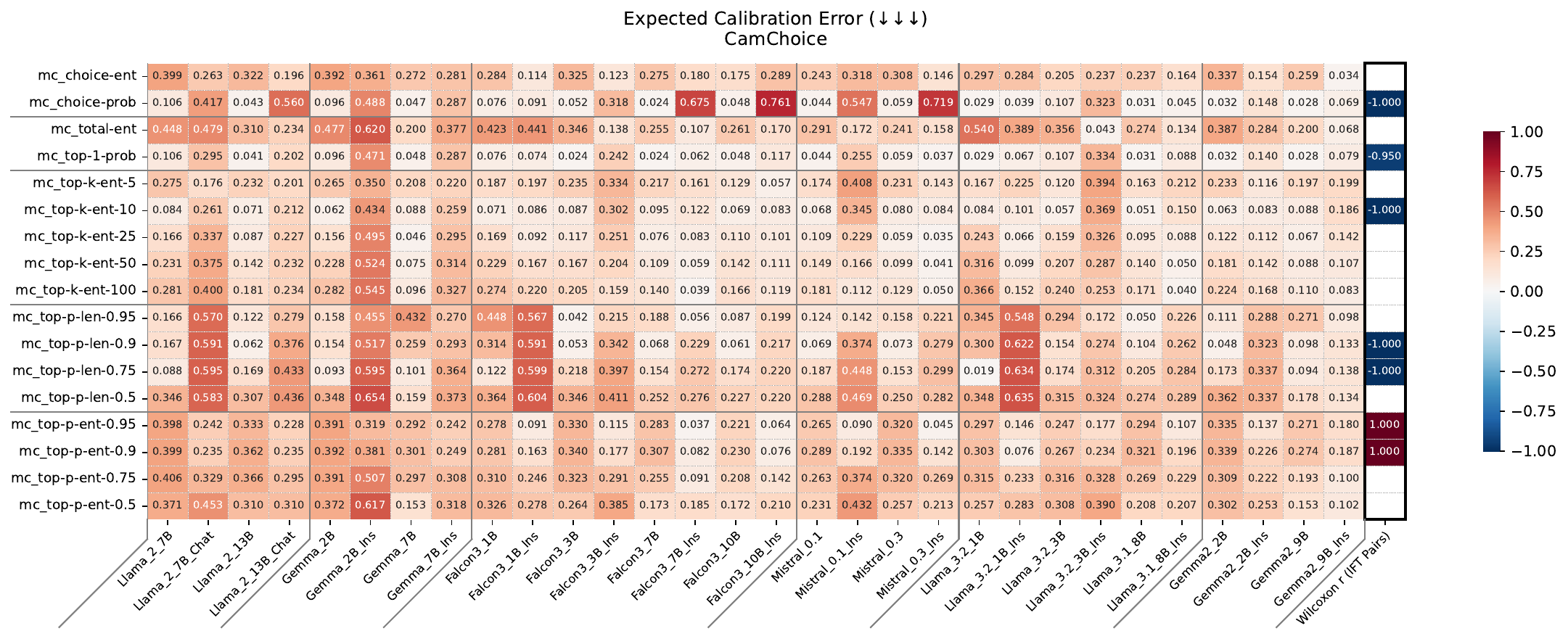}
    \caption{ECESweep results for the CamChoice dataset. Darker cells indicate higher calibration error and thus lower calibration. The rightmost column shows the Wilcoxon effect size between each pair of base vs instruct models, with blank cells indicating insignificance. Models are grouped by family, models are sorted within family by model size, and families are sorted by mean per-column maximum correlation}
    \label{fig:app-ece-first}
\end{figure*}
\begin{figure*}
    \centering
    \includegraphics[width=\linewidth]{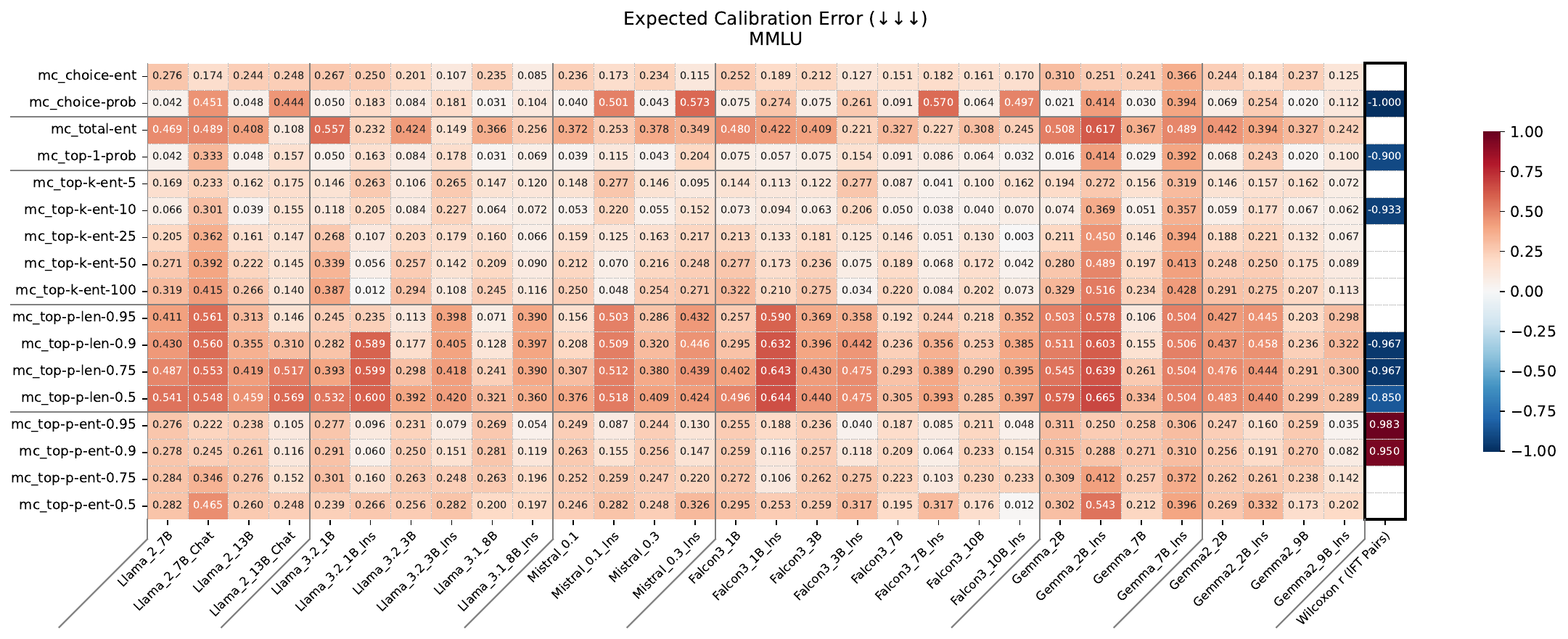}
    \caption{ECESweep results for the MMLU dataset. Darker cells indicate higher calibration error and thus lower calibration. The rightmost column shows the Wilcoxon effect size between each pair of base vs instruct models, with blank cells indicating insignificance. Models are grouped by family, models are sorted within family by model size, and families are sorted by mean per-column maximum correlation}
\end{figure*}
\begin{figure*}
    \centering
    \includegraphics[width=\linewidth]{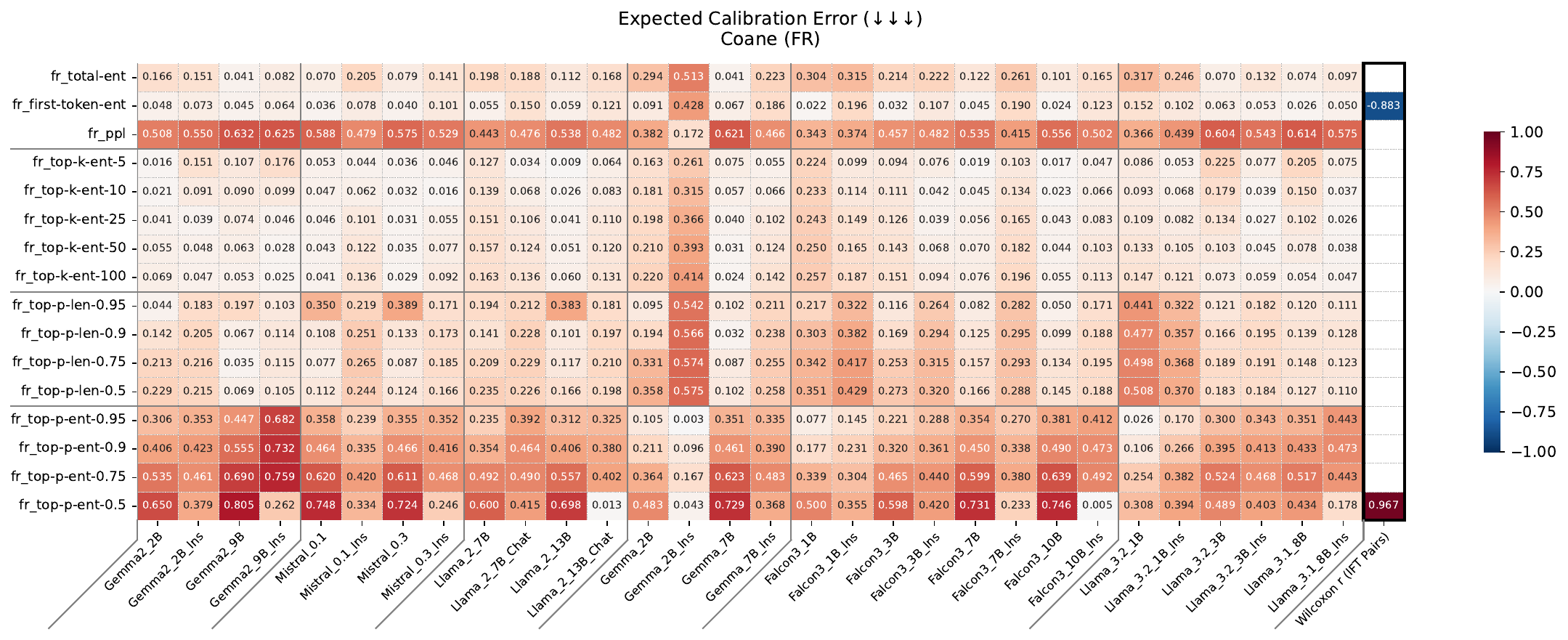}
    \caption{ECESweep results for the FR version of the Coane dataset. Darker cells indicate higher calibration error and thus lower calibration. The rightmost column shows the Wilcoxon effect size between each pair of base vs instruct models, with blank cells indicating insignificance. Models are grouped by family, models are sorted within family by model size, and families are sorted by mean per-column maximum correlation}
\end{figure*}
\begin{figure*}
    \centering
    \includegraphics[width=\linewidth]{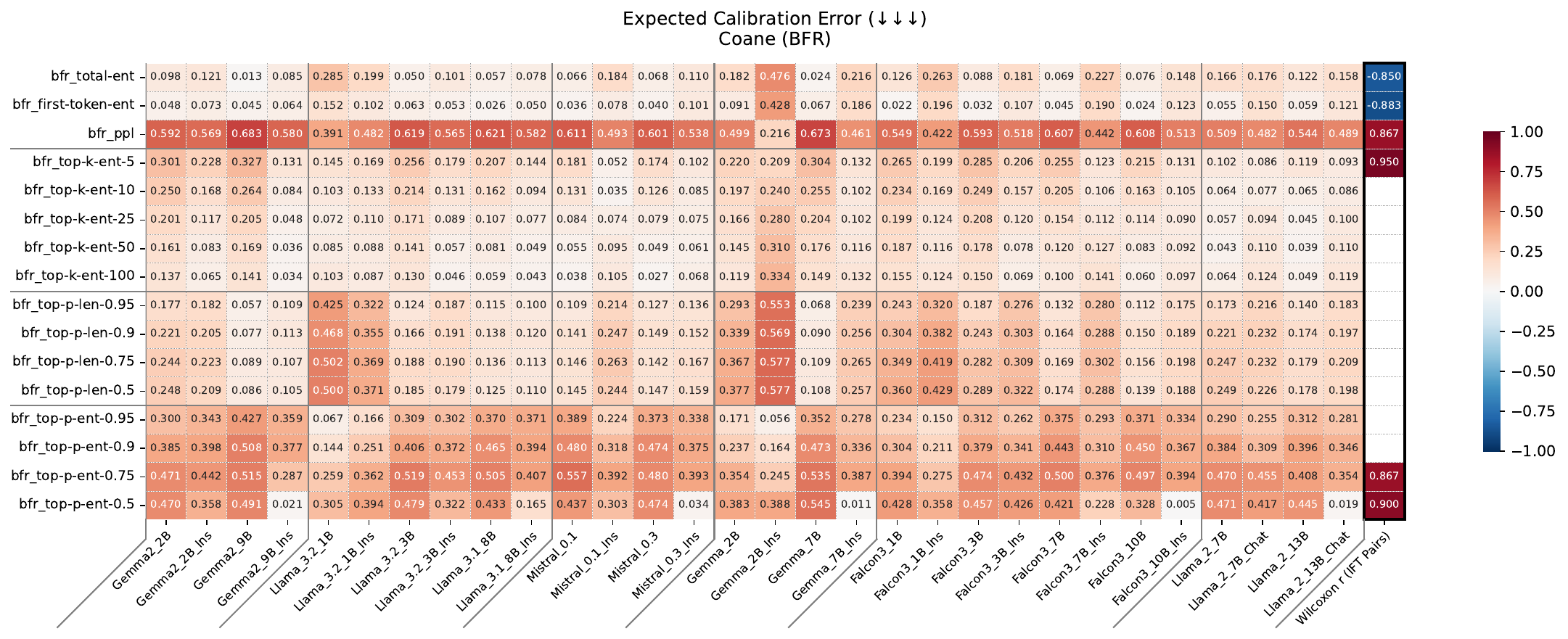}
    \caption{ECESweep results for the BFR version of the Coane dataset. Darker cells indicate higher calibration error and thus lower calibration. The rightmost column shows the Wilcoxon effect size between each pair of base vs instruct models, with blank cells indicating insignificance. Models are grouped by family, models are sorted within family by model size, and families are sorted by mean per-column maximum correlation}
\end{figure*}
\begin{figure*}
    \centering
    \includegraphics[width=\linewidth]{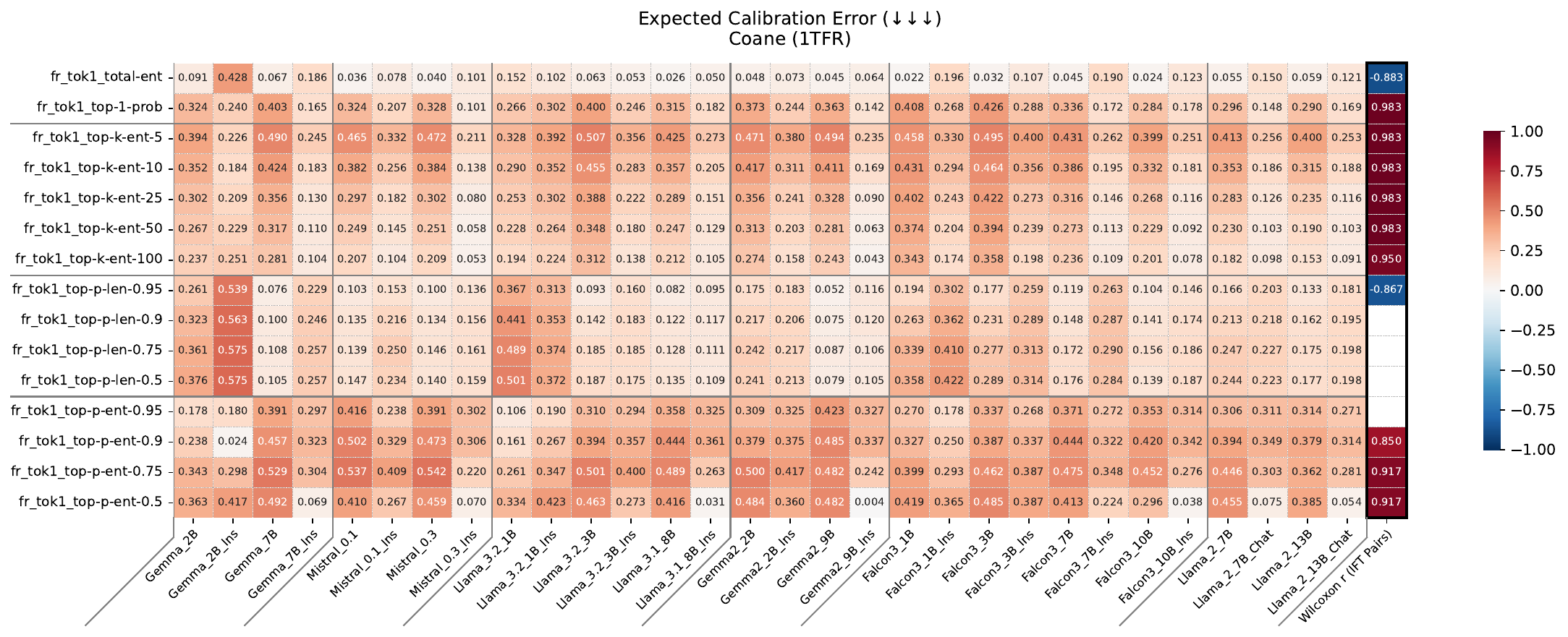}
    \caption{ECESweep results for the 1TFR version of the Coane dataset. Darker cells indicate higher calibration error and thus lower calibration. The rightmost column shows the Wilcoxon effect size between each pair of base vs instruct models, with blank cells indicating insignificance. Models are grouped by family, models are sorted within family by model size, and families are sorted by mean per-column maximum correlation}
    \label{fig:app-ece-last}
\end{figure*}

\begin{figure*}
    \centering
    \includegraphics[width=\linewidth]{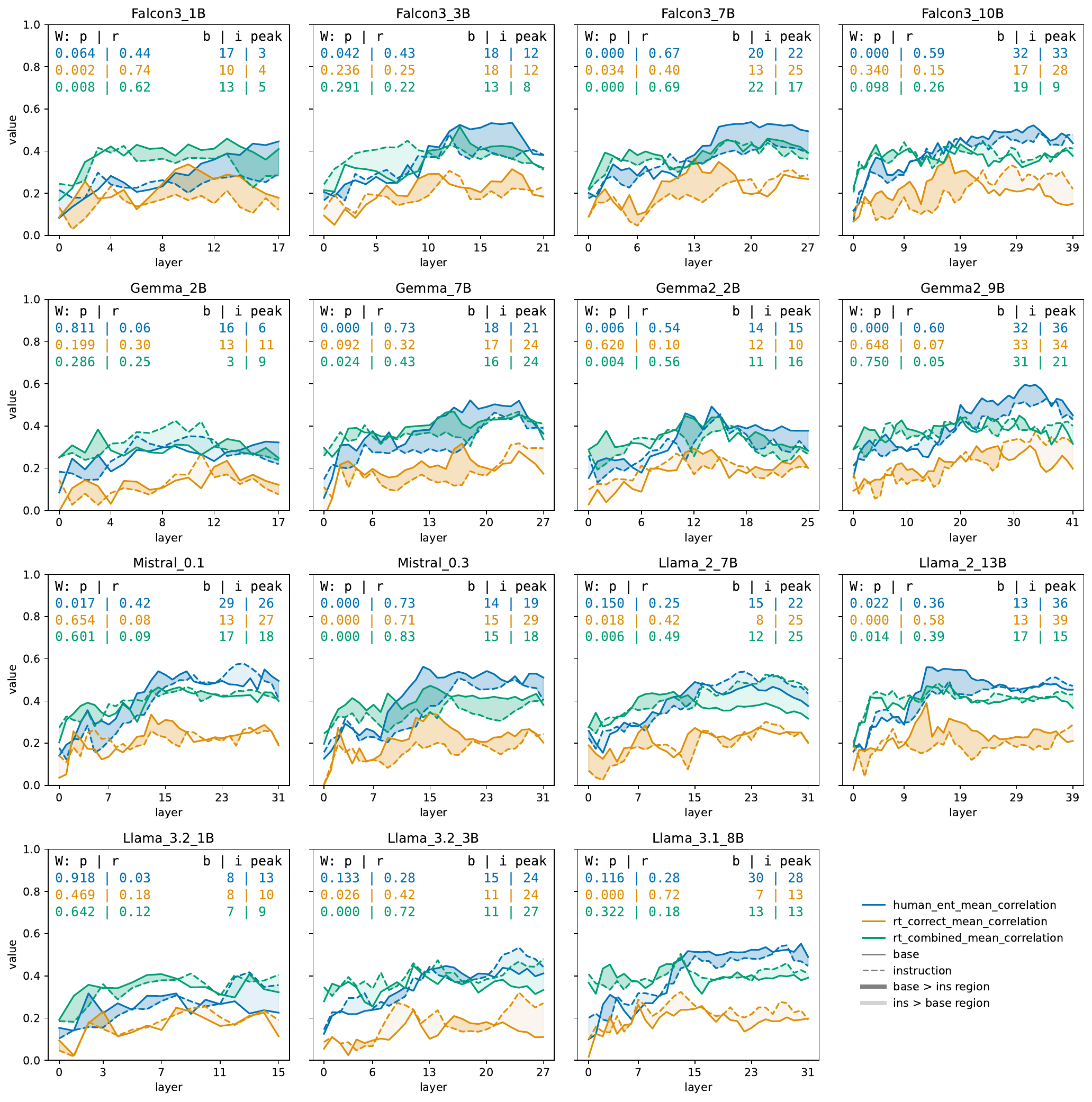}
    \caption{Per-layer model activation probing correlations. Colors indicate the human uncertainty type targeted during linear regression. Shaded regions indicate the gap between the base mode (solid line) and the instruct finetuned variant (dashed line). On-figure annotations in the top-left indicate the p-value and effect size for a Wilcoxon rank sum test between the base correlation and ins correlation across layers. Annotations in the top-right display the layer at which maximum human alignment was measured}
    \label{fig:app-probing}
\end{figure*}

\end{document}